# SUPAID: A Rule mining based method for automatic rollout decision aid for supervisors in fleet management systems.

**Sahil Manchanda**
Conduent Labs
Email: sahilm1992@gmail.com

**Arun Rajkumar**
Conduent Labs
 Email: arun.rajkumar@conduent.com

**Simarjot Kaur**
Conduent Labs
Email: simarjot.kaur@conduent.com

**Narayanan Unny**
Conduent Labs
Email: narayanan.unny@conduent.com

**ABSTRACT**

The decision to rollout a vehicle is critical to fleet management companies as wrong decisions can lead to additional cost of maintenance and failures during journey. With the availability of large amount of data and advancement of machine learning techniques, the rollout decisions of a supervisor can be effectively automated and the mistakes in decisions made by the supervisor learnt. In this paper, we propose a novel learning algorithm SUPAID which under a natural 'one-way efficiency' assumption on the supervisor, uses a rule mining approach to rank the vehicles based on their roll-out feasibility thus helping prevent the supervisor from making erroneous decisions. Our experimental results on real data from a public transit agency from a city in U.S show that the proposed method SUPAID can result in significant cost savings.

INTRODUCTION

Deciding whether a vehicle should be rolled for a revenue trip is a key issue for companies (public or private) with a fleet of vehicles. A wrong decision might incur not just additional cost of maintenance but can lead to incidents such as failure of parts during the journey. In these companies, such rollout decisions are usually made manually by a supervisor, who by looking at the list of pending tasks (issues) and her experience, makes a decision. With the advent of machine learning and data mining techniques, it is natural to consider automating the roll-out decision by learning from past data. However, applying a predictive machine learning algorithm based on past decision of the supervisor has an important but often overlooked drawback. Such algorithms end up *modelling the supervisor* as the data available is based on her past decisions. Thus, black box machine learning algorithms can only hope to be as good as the supervisor. Specifically, these algorithms might fail to capture human errors in judgement or general inefficiency of a supervisor (due to lack of experience etc).

In this paper, we propose a novel decision support system to recommend roll outs where we not learn to mimic the supervisor but also determine situations where the supervisor could have been wrong in her judgement and suggest corrective actions. By automating the decision making process, we not only save time in making the decision but also make more informed and hence less costly decisions. Our method assumes that the supervisor is *one-way efficient.* We define a one-way efficient supervisor as one who is always correct when she makes a 'no roll out' decision but can go wrong when making a 'roll-out' decision. This is a natural assumption to make in practice as one can expect the supervisor to typically have a good reason to prevent a vehicle from rolling out but might miss out on certain key reasons and hence make a wrong decision to roll out (due to lack of experience, error in judgement etc).

We propose an FP growth *(1)* based frequent pattern algorithm called SUPAID to mine task patterns which are frequent in the dataset and have led to failures when a supervisor makes a roll out decision. These patterns are used to assign a score for each vehicle when rollout decision has to be made. The supervisor can use these scores in making a decision for rollout of vehicles.

RELATED WORK

In this section, we review previous work that is related to the problem in hand. We look at three related areas of work: (1) Fleet selection problem, (2) Decision support system for transit fleet management, (3) Association rule mining applications for transit. The presented literature review is not intended to be comprehensive, but to underline some of the general trends in fleet management for transit operations.

**Fleet Selection Problem**
Vehicle dispatch and scheduling is arguably one of the most important aspects of fleet management. An efficient dispatch scheme that chooses the right vehicle at the right time can improve the fleet performance significantly. In the current scenario, fleet selection is a manual procedure carried out by a fleet expert who identifies the most suitable vehicles that can be sent for dispatch based on the current fleet status. Hadden et al. (*2*) presented a system and method that uses sensor data and fault prediction to determine the potential effects that a degraded system may have on the overall capabilities of a machine, computes health capability score for a machine based on these potential effects and assigns the mission according to machine mission capability. Roddy et al. (*3*) presented a system and method that uses historical operational failure data of a vehicle to calculate the likelihood of failure which is further used to generate a health score for each vehicle and finally identify a set of candidate vehicles to be used for a task. Note that our setting is different from both these methods as we try to learn the mistakes of a supervisor while the previous approaches did not take the supervisor's fallibility into account.

**Decision support system for transit fleet management**

Decision support systems for transit fleet management have been widely studied before in transit planning and fleet management operations. Couillard et al. (*4*) developed a decision support system that can be used by fleet managers to plan several transit activities like determining the optimal fleet size/mix and forecasting the demand. Ruiz et al. (*5*) developed a decision support system for a real vehicle routing problem to select the optimum routes from a set of generated feasible routes. Andersson et al. (*6*) developed decision support tools for complex operational management tasks include dispatching, supervising and reconstructing the vehicle schedule to assist the controller in taking such decisions in real time. Although the related problems of vehicle dispatch, scheduling and routing have been widely discussed in the literature and several decision support systems have been developed to provide support for such problems, the problem of optimal fleet selection for dispatch has not received major attention. This is perhaps surprising, given the positive impact of an efficient fleet selection scheme on reducing operational failures, downtime, maintenance cost and maintaining the overall health of the fleet.

**Association rule mining applications for transit**

Data mining techniques are widely used in transportation domain to extract implicit, previously unknown and useful patterns from the data to aid the decision-making process. Kumar et al. (*7*) used data mining techniques on road accidents data to identify the conditions in which an accident is likely to occur. Lee et al. (*8*) presented a knowledge based real-time travel time prediction model which uses data mining techniques to extract useful traffic patterns from the otherwise raw data generated by location-based services and convert them into travel time prediction rules. Li et al. (*9*) proposed a novel evolutionary paradigm named GNP with Estimation of Distribution Algorithms (GNP-EDAs) to solve traffic prediction problems using class association rule mining.

**PROBLEM SETTING**

We work in a real-world fleet management setting wherein for each day, a list of vehicles (say buses) with their corresponding tasks/defects is shown to a supervisor. The supervisor, using her experience makes a decision to roll out some of these buses and decides to send the other buses for maintenance. However, the supervisor might make a wrong decision to roll out a certain bus which might end up with more defects and in turn an increased cost of maintenance. The goal is to develop an algorithm which by analyzing the past data, comes up with a score for each vehicle indicating the suitability for roll out. This score can then help the supervisor make her decisions.

We introduce a natural assumption on the supervisor's capability which we term as 'one-way efficient' supervisor. A one-way efficient supervisor is one who does not make a mistake on a 'no roll-out' decision but might potentially be wrong when making a 'roll-out' decision. This is natural as the supervisor typically has a good reason to reject a bus from rolling out but may have missed important factors while making a roll-out decision.

**Background**

We begin by introducing the necessary background definitions for our algorithm in Table 1. To follow the definitions with an example, refer to Table 2.

**Item-set Mining**

Let $I = \{d_1, d_2, d_3, d_4\}$ be set of attributes called items (tasks/defects). Let $T = \{T_1, T_2, T_3\}$ be set of transactions (bus with a set of defects) where each $T_i$ contains a subset of items in $I$.

Given a list of transactions $T$, the goal is to find all common sets of items defined as those items which exist at least $K$ (called as support) times in our transactions. This is called frequent item-set mining *(10)*. Popular item-set mining algorithms are FP-growth *(1)* and Apriori *(11)*. An important term regarding to item-set mining is support of an item-set. Support for item-set is the number of times an item-set has appeared in transactions.

Example: Let $T_1 = \{d_1, d_2\}$, $T_2 = \{d_2, d_3\}$ and $T_3 = \{d_1, d_3\}$. In this example, some item-sets are $\{d_3\}$, $\{d_1\}$ and $\{d_1, d_2\}$. In this case, support of $\{d_1\}$ is 2 since it appears in transaction $T_1$ and $T_3$. Similarly support of $\{d_1, d_2\}$ is 1 since it appears only in one transaction $T_1$.

**TABLE 1 Definition**

| | |
|---|---|
| [N] | Set $\{1, 2, \ldots, N\}$ |
| *Transactions* | A list of all vehicles with their defect states across multiple days. These are sorted by vehicle id and for each vehicle, sorted in ascending order of timestamp as seen in Table 2. |
| *Support(item_set)* | Function that returns the number of times *item_set* has appeared in defect state column of *Transactions*. For example $\{d_1\}$ item-set appears 5 times in defect state column of transactions as shown in Table 2. |
| *DefState* | Defect state of a vehicle is the set of unresolved defects present in a vehicle. For example, if defects $d_1, d_2, d_3, d_4$ are present in a vehicle *V* then *DefState* is denoted as $\{d_1, d_2, d_3, d_4\}$. |
| *Defects (v)* | Function which returns Defect State at index *v* in *Transactions*. |
| *Available, Down* | The Vehicle Status is marked as *Available* or *Down* based on supervisor's decision |
| *VehicleStatus(i)* | Function which returns the Vehicle Status at $i^{th}$ index of Transactions. Example: *VehicleStatus* (5) returns Available as per Table 2. |
| *VehicleId(i)* | Function that returns vehicle id at $i^{th}$ index of Transactions. |
| *Remains available* | A transition for a vehicle from day *'D'* to day *'D+1'* is marked as *Remains* |
|                   | *available* if the status of the vehicle on day '*D*' is *Available* and day '*D+1*' is *Available*.                                                                                                                                                                                                                                                                     |
| Bad defect state  | *DefState* of vehicle is labelled as *Bad defect state* if the particular *DefState* across all Transactions is labelled as *Down* by supervisor at least some percentage of time based on a configurable threshold $\delta$. In our case the threshold is set to 85%. For example, if defect state of vehicle is $\{d_0, d_1\}$ and in the dataset $\{d_0, d_1\}$ is marked as *Down* for 20 times and *Available* for 2 times then $\{d_0, d_1\}$ is considered as as *Bad defect state* since $\frac{20 \times 100}{20+2} = 91\%$ which is greater than the threshold 85%. |
| $\delta$          | Configurable threshold for labelling a state as *Bad defect state*.                                                                                                                                                                                                                                                                                                      |
| B                 | Set of all bad defect states.                                                                                                                                                                                                                                                                                                                                            |
| Becomes down      | A transition for a vehicle from day '*D*' to day '*D+1*' is marked as *Becomes down* if the status of the vehicle on day '*D*' is *Available* but on day '*D +1*' is *Down* and the *DefState* of the vehicle on day '*D +1*' is a subset of *B*.                                                                                                                         |
| $\alpha$          | Multiset of defect states that are labelled as *Remains available*. Example:- defect state $\{d_1, d_2, d_3\}$ in Table 2 is an element of this multiset.                                                                                                                                                                                                                 |
| $\beta$           | Multiset of defect states that are labelled as *Becomes down*. Example: - defect state $\{d_6, d_7, d_8\}$ in Table 2 is an element of this multiset.                                                                                                                                                                                                                     |

By using the above definitions, the below preprocessing steps define how we create a labelled data set.

**Preprocessing**
We run a preprocessing procedure to convert the data into the appropriate format as required by the algorithm.

Table 2 shows a snapshot of the real data that captures the changing defect state and availability status of vehicles with time. The data is sorted by Vehicle Id and all observations of a vehicle are sorted by Timestamp. To understand the impact of a roll out decision on the health and availability of a vehicle, we organize the data as a time series per vehicle and analyze the time series to extract defect states that worsened if the vehicle is rolled out with these defects.

TABLE 2: Snapshot of subset of data

| Index | Vehicle Id | TimeStamp of observation | Defect State | Vehicle Status |
|---|---|---|---|---|
| 1 | 1 | 1 Jan 2017 | $d_1, d_2, d_3$ | Available |
| 2 | 1 | 2 Jan 2017 | $d_1, d_2, d_3, d_4$ | Available |
| 3 | 1 | 3 Jan 2017 | $d_1, d_2, d_3, d_4, d_6$ | Down |
| 4 | 1 | 4 Jan 2017 | $d_1, d_2$ | Available |
| 5 | 1 | 5 Jan 2017 | $d_1, d_2$ | Available |
| 6 | 2 | 12 March 2017 | $d_6, d_7$ | Available |
| 7 | 2 | 14 March 2017 | $d_6, d_7, d_8$ | Available |
| 8 | 2 | 15 March 2017 | $d_6, d_7, d_8, d_9$ | Down |

Since we need to understand the impact of rolling out vehicles, we find the next Vehicle Status after a vehicle was run with a particular defect state.

We create two multisets $\alpha$ and $\beta$ where $\alpha$ will consist of defect states which had *Remains available* transition and $\beta$ will consist of defect states which had *Becomes down* transitions. A defect state can appear in both the multisets multiple times.

To create multi sets $\alpha$ and $\beta$, we first create Indices$_\alpha$ and Indices$_\beta$ using Equation 1 and 2 respectively.

$$\begin{aligned} Indices_\alpha = \{i \in [N] \mid &VehicleStatus(i) = Available \\ &\& VehicleStatus(i+1) = Available \\ &\& VehicleId(i) = VehicleId(i+1) \} \end{aligned} \quad (1)$$

$$\begin{aligned} Indices_\beta = \{i \in [N] \mid &VehicleStatus(i) = Available \\ &\& VehicleStatus(i+1) = Down \\ &\& Defects(i+1) \in B \\ &\& VehicleId(i) = VehicleId(i+1) \} \end{aligned} \quad (2)$$

The multisets $\alpha$ and $\beta$ are defined using Equation 3 and 4.

$$\alpha = \{ Defects(v) \mid v \in Indices_\alpha \} \quad (3)$$

$$\beta = \{ Defects(v) \mid v \in Indices_\beta \} \quad (4)$$

The labelled data we obtain would look like as in Table 3.
For data presented in Table 2, $\{d_6, d_7, d_8\}$ and $\{d_1, d_2, d_3, d_4\}$ will be members of $\beta$ and $\{d_1, d_2, d_3\}$, $\{d_6, d_7\}$, $\{d_6, d_7\}$, $\{d_1, d_2\}$ and $\{d_1, d_2\}$ will be members of $\alpha$ multiset.

TABLE 3: Labelled data

| Index | Vehicle Id | TimeStamp | Defect State | Current Status | Next Status | Label |
|---|---|---|---|---|---|---|
| 1 | 1 | 1 Jan 2017 | $d_1, d_2, d_3$ | Available | Available | Remains available |
| 2 | 1 | 2 Jan 2017 | $d_1, d_2, d_3, d_4$ | Available | Down | Becomes down |
| 3 | 1 | 4 Jan 2017 | $d_1, d_2$ | Available | Available | Remains available |
| 4 | 1 | 5 Jan 2017 | $d_1, d_2$ | Available | Available | Remains available |
| 5 | 2 | 12 March 2017 | $d_6, d_7$ | Available | Available | Remains available |
| 6 | 2 | 13 March 2017 | $d_6, d_7$ | Available | Available | Remains available |
| 7 | 2 | 14 March 2017 | $d_6, d_7, d_8$ | Available | Down | Becomes down |

**ALGORITHM**

In this section, we present SUPAID, an algorithm that generates suitability scores for each vehicle in a fleet to assist the supervisor in making the rollout decisions for a day. The algorithm runs in three phases, a training phase followed by a prediction phase and finally a ranking phase.

**Step 1:**
**Training Phase**
The training phase is a one-time process that uses the historical decisions made by the supervisor to extract frequently occurring defect patterns (item sets) and estimate their impact on a vehicle's future availability. This is done by computing a quantity called 'item set ratio' which we define below. The exact procedure in the training phase is summarized in the flowchart below in Figure 1.

**Mining item-sets**
The algorithm starts by finding item sets in $\beta$ (multiset of defect states which were part of Becomes down transitions). FP-growth (1) item-set mining algorithm is used to compute the item sets of $\beta$. Let the mined item-sets be called as $IS_{bd}$. Now in order to calculate the likelihood of failure when a particular item set is present, we compute the ratio of frequency of each item set appearing in $\beta$ to frequency of that item set in $\alpha$. This ratio is termed as *item set ratio (ISR)*. In order to do this first we compute support of each item set $Iset \in IS_{bd}$ in both $\alpha$ and $\beta$ multisets using Equation 5.

$$Support(Iset, T) = |\{t \in T; Iset \subseteq t\}| \qquad (5)$$

Here $T$ can be $\alpha$ or $\beta$.
Further, once the support for each item set $Iset \in IS_{bd}$ is computed, the item set ratio for each item set $Iset \in IS_{bd}$ is computed using Equation 6.

$$ISR\,[Iset] = Support(Iset, \beta) / Support(Iset, \alpha) \qquad (6)$$
$$\forall\, Iset \in IS_{bd}$$

The ratio for each item set is stored in dictionary named as *ISR* where key of dictionary is item set and value is the ratio. This dictionary can be further utilised to calculate score for a defect state of a vehicle.

Note that for a given item set, the item set ratio represents the likelihood of the item set causing a down state in the future as opposed to an available state. Therefore, a high item set ratio indicates a higher probability of the defect combination resulting in a breakdown. It is important to note that, by including item sets of all sizes in the frequent item set mining, we are also able to model the dependencies and the joint impact of multiple defects on the health and availability of a vehicle. Also, to reduce the computational overhead, the support and the item set ratio is computed only for item sets in $\beta$ multiset rather than the entire item sets in the dataset.

Let us take an example to show how item sets ratio is computed for each item set. As per Table 3, $\{d_6, d_7\}$ has support of 2 in $\alpha$ (*Remains available* multiset) since it is labelled as *Remains available* twice (index 5 and 6) and it has support of 1 in $\beta$ multiset since it is labelled *Becomes down* only once (index 7). Using Equation 6, we get $ISR\,[\{d_6, d_7\}] = 2$.

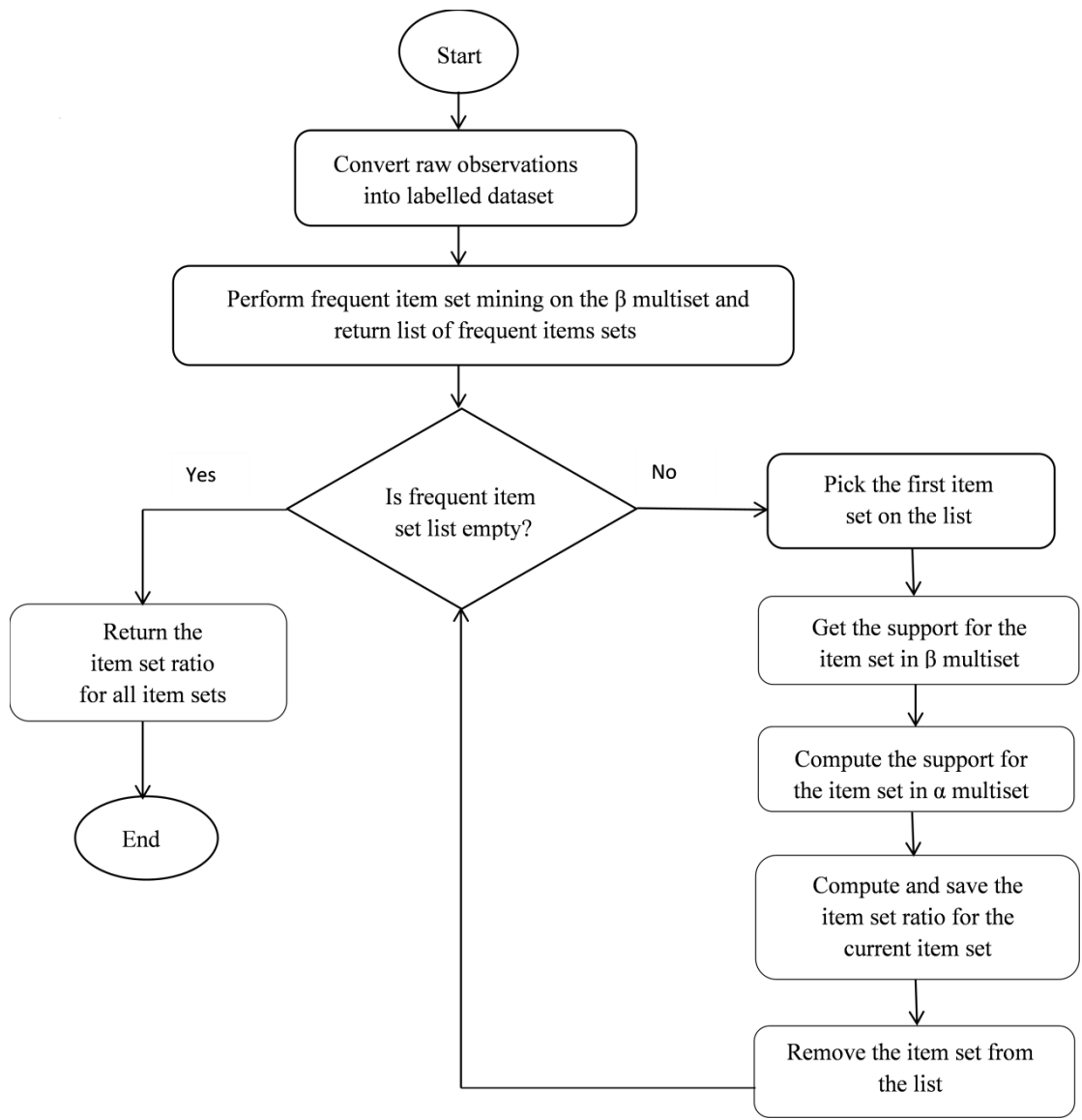

**FIGURE 1** Steps for computing item sets ratio.

**Step2:**
**Prediction Phase**
In the prediction phase, the algorithm accepts as an input, the list of vehicles in the fleet along with the unresolved defects for each vehicle. For each vehicle, given its defect state, the algorithm generates all possible subsets of the defect state as item sets, retrieves the item set ratio for all the generated subsets as computed in the training phase, and returns the maximum item set ratio as the suitability score for the vehicle. A high score therefore indicates a higher likelihood of the vehicle entering a down state if it is rolled out with the current defect state. The prediction phase procedure is summarized in the flowchart in Figure 2.

In order to calculate the score of a particular defect state *DefState* of vehicle, we use the item-sets ratio *ISR* which was calculated in previous step. We need to find subset of *DefState* of vehicle which has the highest item-set ratio. Greater the item-set ratio, greater is the likelihood of failure. Since it is not computationally feasible to calculate exponentially large number of subsets of *DefState* so we take intersection of *DefState* with all item-sets mined in Step 1 and find the item-set which is a subset of *DefState* and has highest item-set ratio.

For given *DefState*, score is calculated using Equation 7.

$$Score = \max_{Iset \in IS_{bd}} ISR[Iset \cap DefState] \tag{7}$$

where $ISR[\emptyset] = 0$.

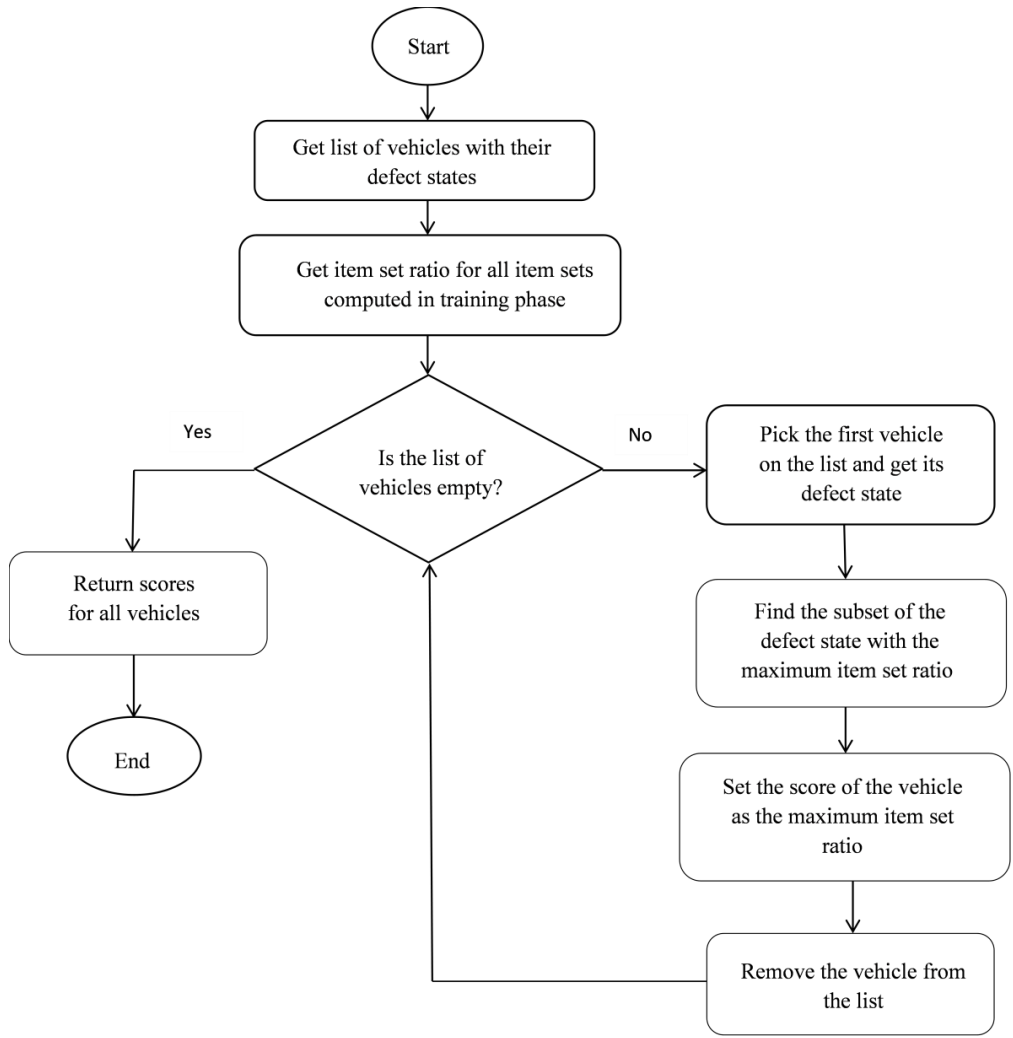

**FIGURE 2** Steps for getting scores for each vehicle depending upon its defect state.

It is clear that based on the scores computed for each vehicle, the algorithm can be used to provide a ranking of the vehicles in the ascending order of their scores. The supervisor can then use this information to select the optimal set from the candidate set of vehicles for a rollout. This is precisely what is done in step 3.

**Step3:**
**Ranking**

Given a list of vehicles each with their *DefState,* the motive is to generate a ranking for all the vehicles according to their *DefState.* Let *VehScore* be defined as a dictionary where key is the vehicle id and value is the score of vehicle as computed by SUPAID.

The vehicle score dictionary *VehScore* is computed using Step 2 by using *DefState* of each vehicle. The dictionary is sorted in ascending order of score. Higher the score, greater is the likelihood of failure and vice-versa. The vehicles in beginning of list are more suitable to be rolled out. Figure 2 shows how ranking for vehicles is generated based upon their defect states.

Figure 3 shows an example with three vehicles and how scores and ranks are calculated for each vehicle based upon their current defect states.

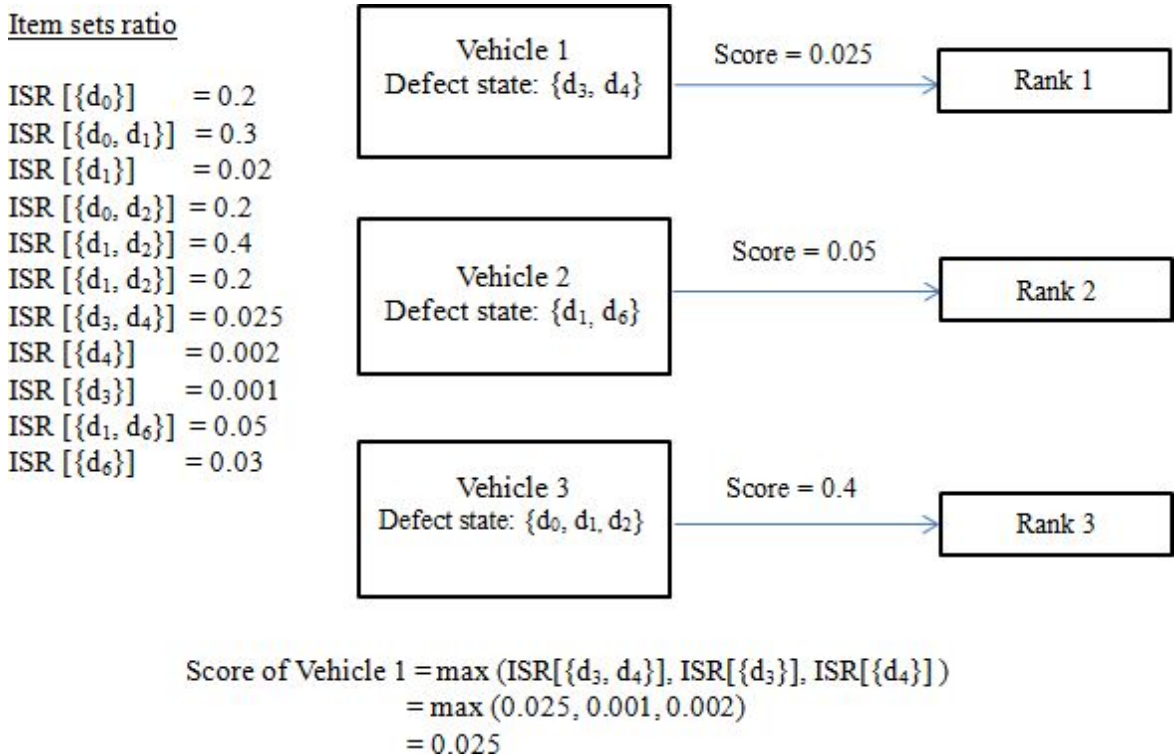

**FIGURE 3** Example of calculation of ranking of a set of three vehicles with their defect states.

# EXPERIMENTS

**Setup:**

Our data is collected from a Public Fleet management company from 25 Dec 2015 to 28 Dec 2016. The data consists of defects state of 945 defects for 249 vehicles for each day in the above time period. We divided our data into 2 parts: training and testing set. Training Data: 25 Dec 2015 to 30 Oct 2016 and Test Data: 02 Nov 2016 to 28 Dec 2016. The training data was used to calculate item sets as done in step 1 of the SUPAID algorithm. For each day in testing data set, we find out how many rollouts were made. We run our algorithm by giving input of the defect state of all vehicles on that day (which were rolled out or not rolled out by the supervisor) and generate a ranking based upon generated scores as described in Ranking step of SUPAID. Only the top N vehicles are selected where N is the number of vehicles rolled out by supervisor on a particular day.

**Evaluation Methodology:**

For evaluating our algorithm on rollouts we use testing data set where supervisor has taken decision to rollout vehicles. We use this data in order to calculate the impact of rollouts on the maintenance cost. We exploit the temporal aspect of our data and observe the change in cost of maintenance required for each vehicle after a rollout on a particular day. To calculate cost of rollout we find difference between future and current costs of repair.
For instance, if cost of repair of a vehicle having defects $\{d_1, d_2, d_3, d_4\}$ is $40 and next day after rollout we see defect state as $\{d_1, d_2, d_3, d_5\}$, which has cost of repair as $50 then the cost incurred due to rollout is considered to be $10. We take the average difference of incurred cost in the next $\theta$ days (we have experimented with different values of $\theta$). This will help us in finding the cost which is incurred as a result of rollout.

For vehicles which were not rolled out by supervisor but our algorithm SUPAID recommends it to be rolled out, we use the following procedure to compute cost. For every rollout occurred for every state in the past, we calculate cost of rollout using the same strategy as described earlier i.e. using cost difference of repair between next few days after rollout and current day. If a state has occurred in multiple rollouts then we take median rollout cost for that state. Once we have past rollout costs for all the states, to calculate estimated cost of rollout of a vehicle V (which is not rolled out) with a given defect state, we compare it with all defect states that have occurred in the past and find the most similar one using Jaccard similarity*(12)*. We assign the rollout cost of the most similar state to the present defect state

**Results:**

In this section, the results for SUPAID Algorithm are presented.

We analysed the performance of our algorithm by varying the parameter $\theta$ which controls how many days we consider after rollout to evaluate impact of rollout on cost of maintenance. We experimented $\theta$ as 3 and 4. The results are shown in Table 4.

**TABLE 4: Comparison of total cost of rollout between Supervisor and SUPAID Algorithm.**

| $\theta$ | Cost incurred by Supervisor( in $ ) | Cost incurred by SUPAID( in $ ) |
|---|---|---|
| 3 | 98271.1 | 96657.0 |
| 4 | 115796.2 | 111878.4 |

The best results we obtain are when we use $\theta$ as **4.** For 57 days of test data, the total rollout cost incurred when using Supervisor's decision is **$115796.2**. On the other hand, the cost incurred when using SUPAID Algorithm is **$111878.4**. On an average, per day SUPAID Algorithm incurs **$1962.78** rollout cost compared to **$2031.513** cost incurred when supervisor's decisions are used instead. The SUPAID Algorithm has outperformed the supervisor by **3.38%.**

In order to analyse cost difference per day, we made a comparison of cost incurred due to rollout for 57 days of data. The graph for that experiment is shown in Figure 4 and Figure 5 for different values of $\theta$. In most of the cases, the cost of rollout using SUPAID's recommendation is less compared to Supervisor's rollout cost.

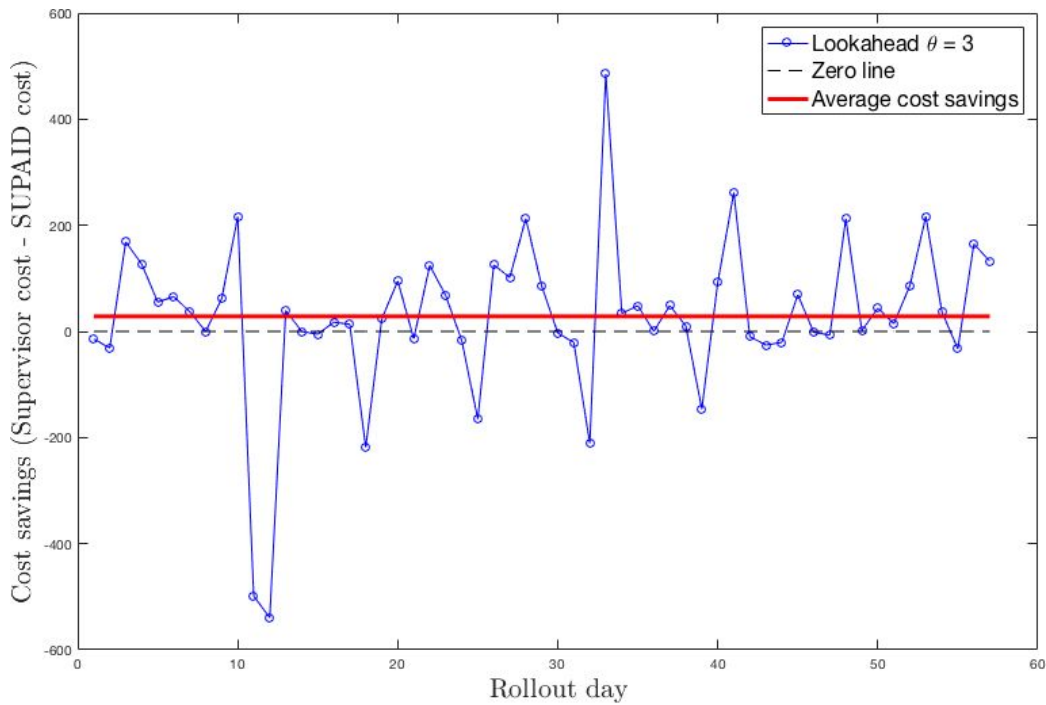

**FIGURE 4 Comparison of cost of rollout using SUPAID's and Supervisor's recommendation with $\theta$=3. Each point shows the difference of cost incurred by supervisor and SUPAID on each day.**

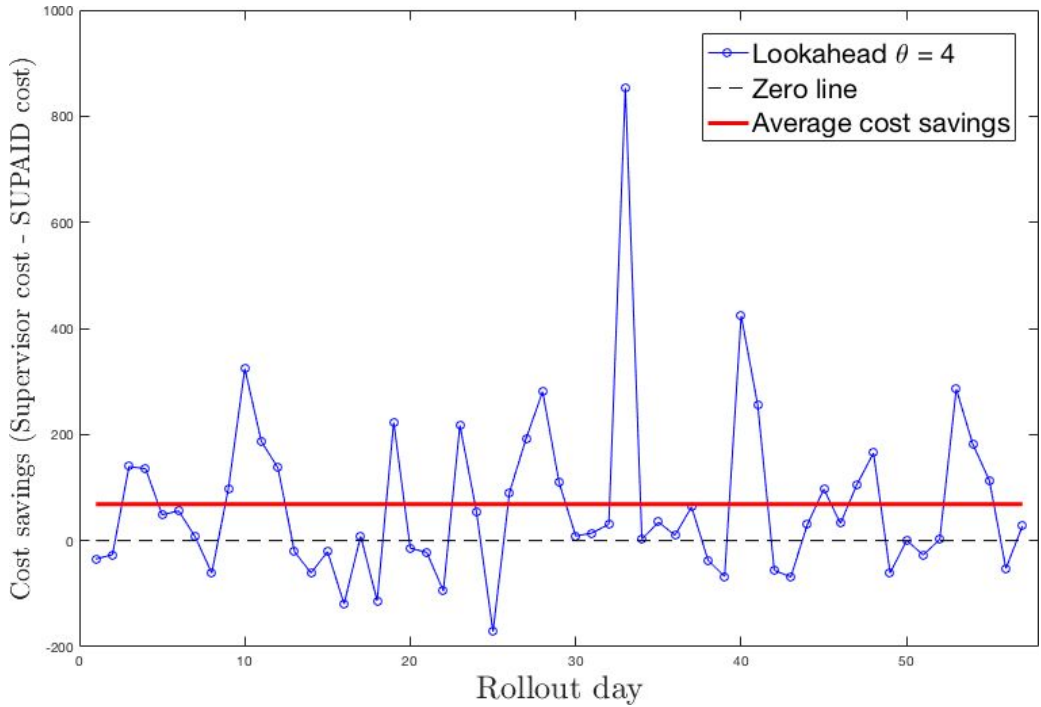

**FIGURE 5** Comparison of cost of rollout using SUPAID's and Supervisor's recommendation with $\theta$ =4. Each point shows the difference of cost incurred by supervisor and SUPAID on each day.

## CONCLUSIONS AND FUTURE WORK

We developed a method SUPAID for automatic roll-out prediction where we not just simply learn to mimic the supervisor but also determine situations where the supervisor could have been wrong in her judgement and suggest corrective actions. With the cost analysis of defects, we have seen that the recommendations of SUPAID were better than that those of the supervisor. Our method has performed better than the supervisor in rolling out vehicles and saving on maintenance cost. Our future work includes using a multi armed bandit strategy to learn to persuade the supervisor in taking the proposed method's decision into account and hence hope to improve the overall cost. Another interesting direction is to explore multi agent reinforcement learning for cost optimisation.